\documentclass{article}
\usepackage{iclr2027_conference,times}
\usepackage{hyperref}
\hypersetup{hidelinks}
\usepackage{url}
\usepackage{booktabs}
\usepackage{multirow}
\usepackage{graphicx}
\usepackage{amsmath,amsfonts}
\usepackage[capitalize]{cleveref}
\crefname{section}{Sec.}{Secs.}
\Crefname{section}{Section}{Sections}
\Crefname{table}{Table}{Tables}
\crefname{table}{Tab.}{Tabs.}
\Crefname{equation}{Equation}{Equations}
\crefname{equation}{Eqn.}{Eqns.}

\title{MeteoVerse: Unified Weather-Controllable \\Video World Model}

\author{
\textbf{Renlong Wu}$^{1}$ \quad
\textbf{Guanqiao Wang}$^{1}$ \quad
\textbf{Xuan Shang}$^{1}$ \quad
\textbf{Yin Hanming}$^{1}$ \\
\textbf{Xiaoxiao Sheng}$^{2}$ \quad
\textbf{Tianyu Huang}$^{2}$ \quad
\textbf{Hui Li}$^{1}$ \quad
\textbf{Wangmeng Zuo}$^{1}$ \\[0.5em]
$^{1}$Harbin Institute of Technology \quad
$^{2}$Huawei
}

\iclrfinalcopy
\begin{document}

\maketitle
\fancyhead{}

\begin{abstract}
Video world models aim to predict future content from an observed scene while following prescribed camera motion. Real-world scene evolution is determined not only by changes in viewpoint and object dynamics, but also by environmental conditions such as weather, which can substantially alter scene appearance and visibility. Modeling such realistic weather evolution is challenging because the required weather modification depends jointly on the observed and desired weather states. Depending on their relation, the model may need to preserve, introduce, or remove a weather effect. Existing video world models typically leave this weather transition implicit, forcing the generation backbone to infer weather evolution together with scene dynamics and camera motion, which leads to imprecise weather control. To address this limitation, we propose MeteoVerse, a unified weather-controllable video world model that generates future videos from a single sunny or adverse-weather image, conditioned on a weather-free scene description, a target-weather instruction, and a camera trajectory. Rather than conditioning only on the desired weather, MeteoVerse explicitly estimates the observed and target weather states and represents the required weather transition. A transition-aware mixture of weather experts (MeteoMoE) then translates this transition into category-specific residual weather features, unifying weather preservation, introduction, and removal while enabling fine-grained control over introduced weather intensity. We further construct the MeteoVerse dataset with over 50K real-world weather video clips, generated sunny counterparts, disentangled scene and weather descriptions, weather-intensity annotations, and camera trajectories. Extensive quantitative and qualitative experiments demonstrate substantially improved weather controllability while retaining competitive scene consistency and camera-control performance. Project page: \url{https://meteoverse.github.io/}.
\end{abstract}

\section{Introduction}
Video world models aim to predict future content from an observed scene while following prescribed camera motion.
While recent advances have enabled increasingly flexible control over scene content and viewpoint, real-world scene evolution is also shaped by environmental conditions.
Weather is a particularly important factor, as changes in rain, snow, and fog can substantially alter scene appearance and visibility over time.
A world model simulating realistic future observations should therefore model not only what is observed and from where, but also under which weather condition the future unfolds.
Accordingly, we study weather-controllable video world modeling, \emph{i.e.}, given a single image captured under sunny or adverse weather, a weather-free scene description, a target-weather instruction, and a camera trajectory, the model generates a future video that preserves scene identity, follows the prescribed camera motion, and realizes the desired weather evolution through weather preservation, introduction, or removal.

\begin{figure*}[t!]
    \centering
    \includegraphics[width=\textwidth]{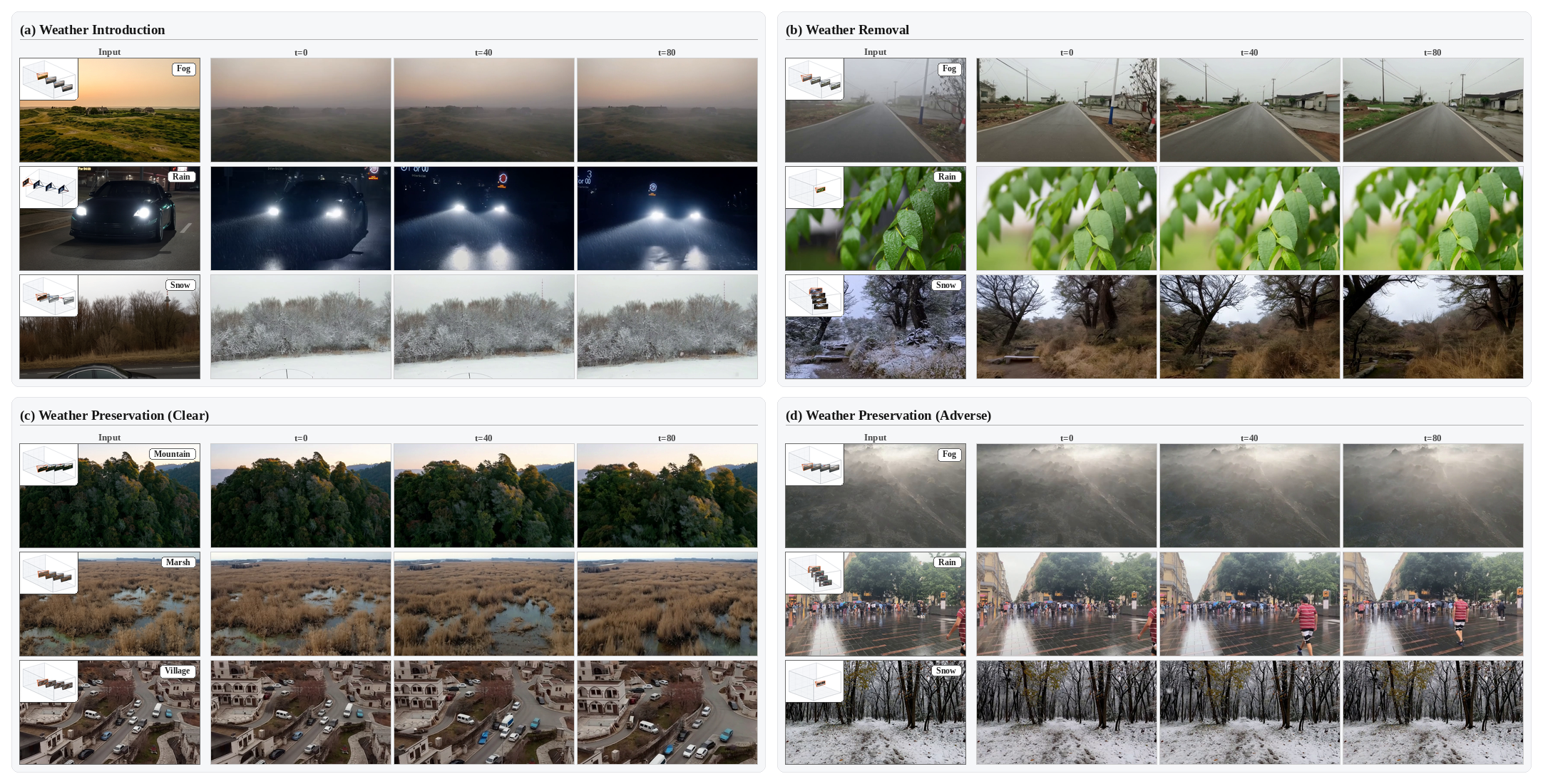}
    \caption{
    Overview of MeteoVerse, which supports weather preservation, introduction, and removal from sunny or adverse-weather inputs under prescribed camera trajectories.
    }
    \label{fig:intro_figure}
    \vspace{-8mm}
\end{figure*}

Existing approaches address weather manipulation under substantially different assumptions.
Reconstruction-based methods can synthesize geometrically consistent weather effects through explicit 3D scene representations or physical simulation, but require multi-view observations and costly per-scene reconstruction~\citep{li2022climatenerf,qian2025weatheredit}.
Video weather editing methods can introduce or remove weather effects while preserving the structure and motion of an existing video~\citep{lin2025weatherweaver,qian2026weathervid}, but assume that the complete source video is already available.
More recently, Holo-World brings weather control into single-image video world modeling by jointly controlling camera motion and adverse-weather generation~\citep{yin2026holoworld}.
However, it assumes a sunny input and mainly focuses on weather introduction, leaving weather removal unexplored.
These limitations motivate a unified formulation that can operate from either sunny or adverse-weather observations and explicitly control how weather evolves into the future.

The key challenge is that a target-weather instruction specifies the desired weather condition, but not the transformation required to reach it.
The same target condition may correspond to fundamentally different operations depending on the weather already present in the input.
For example, a rainy target requires introducing rain from a sunny observation or preserving rain from an already rainy observation, while a sunny target requires removing adverse weather from an adverse-weather observation.
Therefore, effective weather control requires explicitly reasoning about the transition between the observed and target weather states rather than relying on the target condition alone.
When this transition is left implicit, the video backbone must simultaneously infer the observed weather, interpret the target instruction, determine the required modification, and model scene evolution and camera motion.
This entanglement makes precise weather control particularly difficult.

To address this challenge, we propose MeteoVerse, a unified weather-controllable video world model that explicitly models the transition between observed and desired weather states.
Given the input image and target-weather instruction, a weather-state predictor estimates continuous observed and target states over rain, snow, and fog, from which the required weather transition is derived.
Rather than directly conditioning the entire generation backbone on the target weather, we introduce a transition-aware mixture of weather experts (MeteoMoE) to translate the transition into residual weather modifications.
Specifically, category-specific rain, snow, and fog experts extract scene-adaptive weather features, whose responses are modulated according to the direction and magnitude of the corresponding transition components and then fused into a transition-aware representation.
The resulting representation is injected into the video world-model backbone, while the weather-free scene description and camera trajectory provide semantic and viewpoint conditions independently.
This decomposition allows the pretrained backbone to preserve the underlying scene evolution, while MeteoMoE focuses on the weather modification required to reach the target state.
As a result, a single model can support weather preservation, introduction, and removal, together with fine-grained control over introduced weather intensity.

Training such a model requires supervision covering diverse weather transitions.
We therefore construct the MeteoVerse dataset, containing over 50K real-world rain, snow, and fog video clips together with generated sunny counterparts, disentangled scene and weather descriptions, category-specific weather-intensity annotations, and camera trajectories.
Extensive quantitative and qualitative experiments demonstrate that MeteoVerse substantially improves weather controllability over existing controllable video world models, particularly when the desired weather differs from the observed condition, while retaining competitive scene consistency and camera-control performance.

Our contributions are summarized as follows:
\begin{itemize}
    \item We introduce MeteoVerse, a unified weather-controllable video world model that predicts camera-controlled future videos from either sunny or adverse-weather observations and supports weather preservation, introduction, and removal.

    \item We propose explicit weather transition modeling together with MeteoMoE, which translates weather-state changes into category-specific residual weather features and enables fine-grained control over introduced weather intensity.

    \item We construct the MeteoVerse dataset with over 50K real-world weather video clips, generated sunny counterparts, disentangled scene and weather descriptions, weather-intensity annotations, and camera trajectories, providing supervision for diverse weather transitions. Extensive experiments demonstrate substantially improved weather controllability while maintaining competitive scene consistency and camera-control performance.
\end{itemize}

\section{Related Work}
\label{sec:related_work}

\subsection{Video World Models}

Camera-controllable video generation has evolved from motion control to geometry-aware scene exploration.
MotionCtrl, CameraCtrl, and CamI2V introduce explicit camera control into video generation~\citep{wang2024motionctrl,he2024cameractrl,zheng2024cami2v}, while ViewCrafter, ReCamMaster, TrajectoryCrafter, and Voyager further support camera-guided scene exploration or trajectory manipulation~\citep{yu2024viewcrafter,bai2025recammaster,yu2025trajectorycrafter,huang2025voyager}.
Video world models extend this paradigm toward interactive future prediction, including GAIA-1, Genie, GameNGen, Matrix-Game, and LingBot-World~\citep{hu2023gaia1,bruce2024genie,valevski2024gamengen,zhang2025matrixgame,robbyant2026lingbotworld}.
Most closely related, Holo-World introduces weather control into single-image video world modeling~\citep{yin2026holoworld}, but mainly considers adverse-weather generation from sunny inputs.
MeteoVerse instead supports both sunny and adverse-weather observations and unifies weather preservation, introduction, and removal.

\subsection{Weather Synthesis Methods}

Weather manipulation has been studied in restoration, reconstruction-based synthesis, and generative editing.
Multi-weather restoration methods remove rain, snow, and fog from degraded observations~\citep{li2020allinone,valanarasu2022transweather,yang2023viwsnet,yang2024difftta}, while ClimateNeRF, WeatherEdit, and WeatherCity achieve view-consistent weather synthesis through explicit scene representations~\citep{li2022climatenerf,qian2025weatheredit,wu2026weathercity}.
Generative approaches such as IntrinsicWeather, WeatherWeaver, and AutoAWG reduce the need for explicit reconstruction~\citep{zhu2026intrinsicweather,lin2025weatherweaver,hu2026autoawg}, but image editing does not predict unseen views and video editing assumes a complete source video.
MeteoVerse instead predicts future observations from a single image while jointly controlling camera motion and weather evolution.
Existing weather datasets mainly target adverse-weather perception, restoration, or editing~\citep{sakaridis2021acdc,sun2022shift,zhang2023weatherstream,lin2025weatherweaver,zhu2026intrinsicweather,yin2026holoworld}.
MeteoVerse provides over 50K real-world weather clips with generated sunny counterparts, disentangled scene and weather descriptions, fine-grained weather-intensity annotations, and camera trajectories to support weather-controllable video world modeling.

\begin{figure*}[t]
    \centering
    \includegraphics[width=\textwidth]{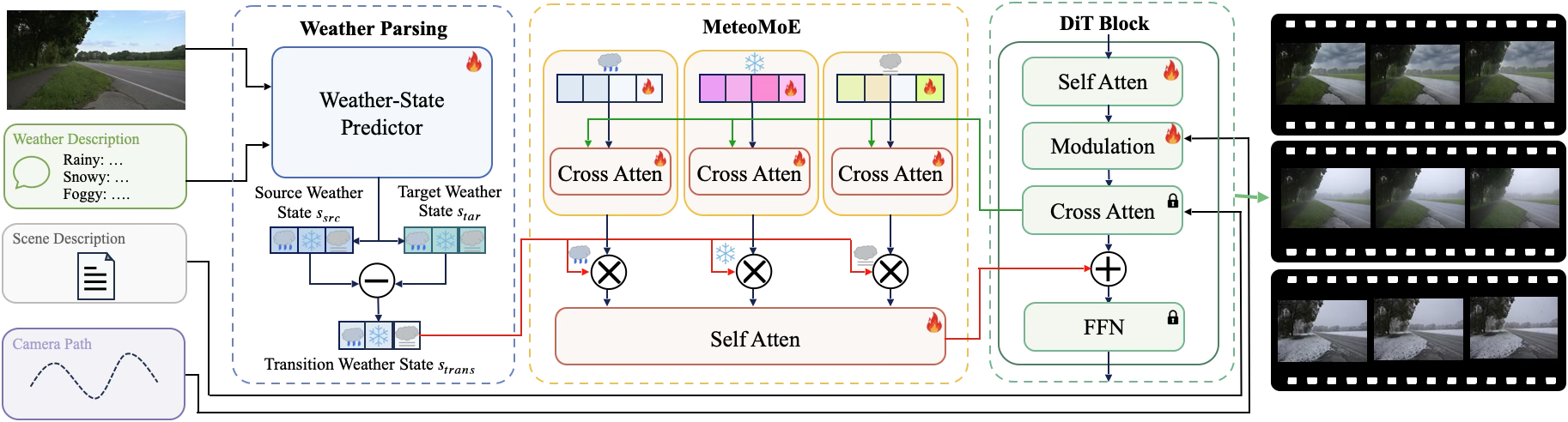}
    \caption{
    Overview of MeteoVerse.
    The weather-state predictor estimates the observed and target weather states and derives the required weather transition.
    MeteoMoE modulates the rain, snow, and fog experts according to the corresponding transition components, fuses their responses through self-attention, and residually injects the resulting transition-aware features into the DiT backbone.
    }
    \label{fig:MeteoVerse}
    \vspace{-4mm}
\end{figure*}

\section{Methods}
\label{sec:method}

\subsection{Problem Formulation and Overview}
\label{sec:method_overview}

Let $I$ denote the current visual observation, $p$ a general text condition, and $\mathcal{C}$ a prescribed camera trajectory.
A controllable video world model with parameters $\theta$ predicts future observations $V_{1:T}$ as,
\begin{equation}
p_{\theta}
\left(
V_{1:T}
\mid
I,
p,
\mathcal{C}
\right).
\label{eq:conventional_wm}
\end{equation}
Although weather information may be implicitly contained in $I$ or $p$, this formulation does not explicitly specify how the observed weather should evolve toward the requested condition.
The video backbone therefore needs to infer the required weather modification together with scene evolution and camera-induced changes, which can lead to imprecise weather control.
As illustrated in \cref{fig:MeteoVerse}, MeteoVerse separates the general text condition into a weather-free scene description $p_{\mathrm{s}}$ and a target-weather instruction $p_{\mathrm{w}}$.
A weather-state predictor $P_{\phi}$ estimates the weather state observed in the input image and the desired target weather state,
\begin{equation}
\left(
\mathbf{s}_{\mathrm{src}},
\mathbf{s}_{\mathrm{tar}}
\right)
=
P_{\phi}
\left(
I,
p_{\mathrm{w}}
\right),
\qquad
\mathbf{s}_{\mathrm{trans}}
=
\mathbf{s}_{\mathrm{tar}}
-
\mathbf{s}_{\mathrm{src}}.
\label{eq:weather_transition_overview}
\end{equation}
Here, $\mathbf{s}_{\mathrm{src}}$ describes the weather condition observed in the input image, while $\mathbf{s}_{\mathrm{tar}}$ represents the desired future weather condition.
Their difference $\mathbf{s}_{\mathrm{trans}}$ characterizes the weather modification required to reach the target state.
We further introduce MeteoMoE, a transition-aware mixture of weather experts that converts $\mathbf{s}_{\mathrm{trans}}$ into residual weather conditioning for the pretrained video backbone.
The resulting generation process can be written as,
\begin{equation}
p_{\theta,\psi}
\left(
V_{1:T}
\mid
I,
p_{\mathrm{s}},
\mathcal{C},
\mathbf{s}_{\mathrm{trans}}
\right),
\label{eq:meteoverse_formulation}
\end{equation}
where $\theta$ denotes the frozen pretrained backbone parameters and $\psi$ denotes the trainable adaptation parameters.
In this formulation, $p_{\mathrm{s}}$ provides scene semantics, $\mathcal{C}$ controls camera motion, and $\mathbf{s}_{\mathrm{trans}}$ specifies the required weather modification.

\subsection{Weather-State Transition Modeling}
\label{sec:weather_state}

We represent each weather state using a continuous intensity vector over rain, snow, and fog, \emph{i.e.},
\begin{equation}
\mathbf{s}
=
\left[
s_{\mathrm{rain}},
s_{\mathrm{snow}},
s_{\mathrm{fog}}
\right]
\in [0,1]^3,
\label{eq:weather_state}
\end{equation}
where each component denotes the intensity of the corresponding weather effect.
Sunny weather is represented by the zero state $\mathbf{s}=[0,0,0]$.
Accordingly, each component $\delta_k$ of $\mathbf{s}_{\mathrm{trans}}$ describes the direction and magnitude of the required modification for weather category $k$.
A positive value indicates that the corresponding weather effect should be introduced.
A negative value indicates that the effect should be removed.
When $\delta_k=0$, no additional modification is required and the observed weather condition is preserved.
We LoRA-fine-tune Qwen3-VL-2B~\cite{bai2025qwen3} as the weather-state predictor $P_{\phi}$.
Its supervision is constructed from the weather annotations described in \cref{sec:dataset_construction}.
The LoRA rank and scaling factor $\alpha$ are set to 64 and 128, respectively.
During video-model training, the annotated weather states are directly used to compute $\mathbf{s}_{\mathrm{trans}}$.
The predictor is frozen and used only at inference time.
For weather introduction from sunny inputs, users may specify $\mathbf{s}_{\mathrm{tar}}$ to control the desired weather intensity.

\subsection{Transition-Aware MeteoMoE}
\label{sec:meteo_moe}

MeteoMoE translates the explicit weather transition $\mathbf{s}_{\mathrm{trans}}$ into residual weather features for the video backbone.
Rain, snow, and fog exhibit distinct visual characteristics, and representing them with a shared weather feature may entangle category-specific effects.
We therefore introduce three weather experts.
The expert for weather category $k$ is represented by a learnable token set $T_k$, where
$k\in\{\mathrm{rain},\mathrm{snow},\mathrm{fog}\}$.
In the $\ell$-th DiT block, let $H_{\mathrm{txt}}^{\ell}$ denote the feature after text cross-attention.
MeteoMoE uses this feature as the query to retrieve category-specific weather information from the expert tokens, \emph{i.e.},
\begin{equation}
E_k^{\ell}
=
\operatorname{CA}_{k}^{\ell}
\left(
\operatorname{Norm}
\left(
H_{\mathrm{txt}}^{\ell}
\right),
T_k,
T_k
\right),
\qquad
k\in
\{
\mathrm{rain},
\mathrm{snow},
\mathrm{fog}
\}.
\label{eq:weather_expert}
\end{equation}
Here, $H_{\mathrm{txt}}^{\ell}$ serves as the query, while $T_k$ provides the keys and values.
This produces a scene-adaptive representation for each weather category.
Each expert response is then modulated by the corresponding transition component $\delta_k$, which are fused through a lightweight self-attention module.
It can be written as,
\begin{equation}
E_{\mathrm{trans}}^{\ell}
=
\operatorname{SA}_{\mathrm{w}}^{\ell}
\left(
\left\{
\delta_k E_k^{\ell}
\right\}_{k\in
\{
\mathrm{rain},
\mathrm{snow},
\mathrm{fog}
\}}
\right).
\label{eq:weather_fusion}
\end{equation}
Finally, $E_{\mathrm{trans}}^{\ell}$ is residually injected into the DiT block as,
\begin{equation}
\widetilde{H}_{\mathrm{txt}}^{\ell}
=
H_{\mathrm{txt}}^{\ell}
+
E_{\mathrm{trans}}^{\ell}.
\label{eq:weather_injection}
\end{equation}
In this way, MeteoMoE focuses on the weather modification specified by $\mathbf{s}_{\mathrm{trans}}$, while the pretrained video backbone retains scene evolution and camera-conditioned future prediction.

\begin{figure*}[t]
    \centering
    \includegraphics[width=\textwidth]{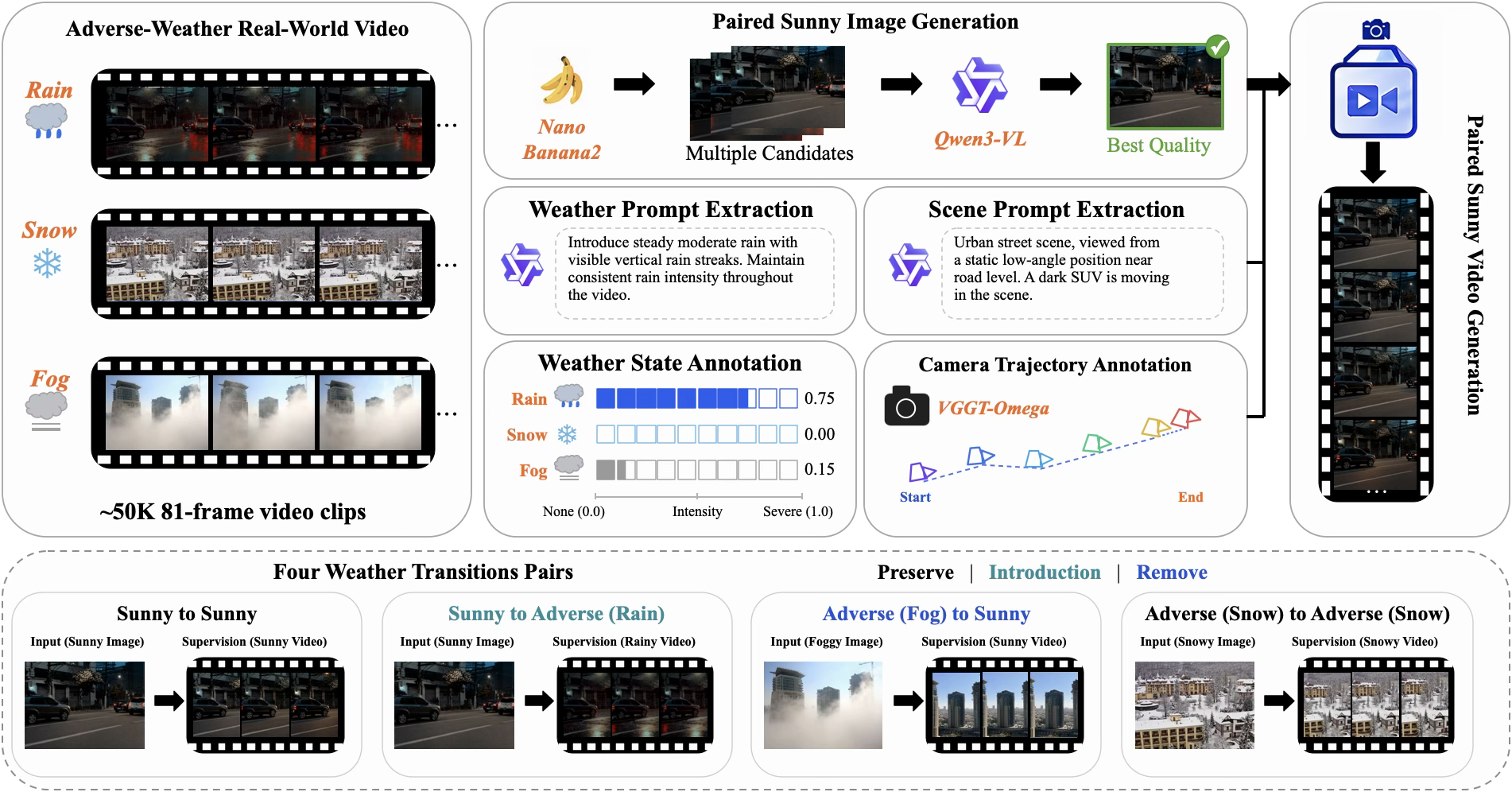}
    \caption{
    Construction of MeteoVerse dataset.
    We collect over 50K 81-frame real-world rain, snow, and fog clips with disentangled scene and weather descriptions, weather-intensity annotations, camera trajectories, and pseudo-paired sunny counterparts.
    The sunny and adverse-weather observations are then recombined to provide supervision for weather preservation, introduction, and removal.
    }
    \label{fig:dataset_construction}
    \vspace{-4mm}
\end{figure*}

\subsection{MeteoVerse Dataset Construction}
\label{sec:dataset_construction}

The construction pipeline is illustrated in \cref{fig:dataset_construction}.
We collect real-world videos with visible rain, snow, or fog and divide them into 81-frame clips, resulting in over 50K adverse-weather clips.
For each clip, Qwen3-VL-8B~\cite{bai2025qwen3} generates a weather-free scene description and a separate weather description.
VGGT-Omega~\cite{wang2026vggt} estimates the camera intrinsics and extrinsics, and clips with severe camera jitter or abrupt pose changes are discarded.
To obtain fine-grained weather states, we adopt a coarse-to-fine intensity annotation strategy.
For each weather category, Qwen3-VL-30B~\cite{bai2025qwen3} first predicts an ordinal intensity level according to visual severity and then estimates a continuous intensity value within the selected level.
Each clip is evaluated with a confidence estimate, and samples with low confidence are removed.
The remaining annotations are further verified by GPT-5.6 Sol~\cite{openai2026gpt56sol}.

\noindent\textbf{Pseudo-paired sunny counterpart generation.}
Since paired sunny and adverse-weather videos of the same real-world scene are difficult to obtain, we construct a pseudo-paired sunny counterpart for each adverse-weather clip.
Given the first adverse-weather frame $I_{\mathrm{adv}}$, Nano Banana 2~\cite{deepmind2026nanobanana2} generates multiple sunny candidates.
Qwen3-VL-30B~\cite{bai2025qwen3} selects the candidate that best removes the adverse-weather effect while preserving scene content.
Samples without a qualified candidate are discarded.
The selected sunny image $I_{\mathrm{sun}}$, together with the weather-free scene description and recovered camera trajectory, is passed to the pretrained LingBot-World~\cite{robbyant2026lingbotworld} to generate the pseudo-paired sunny video $V_{\mathrm{sun}}$.
We discard generated videos with poor visual quality or obvious temporal artifacts, and re-estimate the camera trajectory of the remaining videos using VGGT-Omega~\cite{wang2026vggt} to obtain more accurate annotations of the camera motion.

\noindent\textbf{Training pair construction.}
Let $(I_{\mathrm{adv}},V_{\mathrm{adv}})$ denote the original adverse-weather pair and $(I_{\mathrm{sun}},V_{\mathrm{sun}})$ denote its pseudo-paired sunny counterpart.
We construct four input-target pairs,
\begin{equation}
\mathcal{P}
=
\left\{
(I_{\mathrm{sun}},V_{\mathrm{sun}}),
(I_{\mathrm{adv}},V_{\mathrm{adv}}),
(I_{\mathrm{sun}},V_{\mathrm{adv}}),
(I_{\mathrm{adv}},V_{\mathrm{sun}})
\right\}.
\label{eq:training_pairs}
\end{equation}
The first two pairs provide supervision for sunny and adverse-weather preservation.
The third pair provides supervision for weather introduction, while the fourth provides supervision for weather removal.
Sunny observations are assigned the zero weather state, while adverse-weather observations use their annotated states.
The diverse weather intensities in the collected clips provide supervision for introducing rain, snow, and fog at different intensity levels.

\subsection{Network Architecture and Optimization}
\label{sec:world_model}

We build MeteoVerse upon the pretrained LingBot-World~\cite{robbyant2026lingbotworld}, which consists of a high-noise model for coarse generation and a low-noise model for refinement.
The input image, weather-free scene description, and camera trajectory $\mathcal{C}$ follow the original visual, textual, and camera-conditioning pathways, respectively.
MeteoMoE is inserted after text cross-attention in every DiT block of both stages, with independent parameters for the high-noise and low-noise models.
We freeze the pretrained backbone and optimize MeteoMoE together with LoRA adapters inserted into the self-attention and feed-forward layers.
The LoRA rank and scaling factor $\alpha$ are set to 64 and 128, respectively.

Following LingBot-World, the video model is optimized using a flow-matching objective.
Let $z_t$ denote the latent state at flow timestep $t$ and $v_t$ denote the corresponding target velocity.
The flow-matching loss can be written as,
\begin{equation}
\mathcal{L}_{\mathrm{fm}}
=
\mathbb{E}_{t,z_t}
\left[
\left\|
G_{\theta,\psi}
\left(
z_t,
t,
I,
p_{\mathrm{s}},
\mathcal{C},
\mathbf{s}_{\mathrm{trans}}
\right)
-
v_t
\right\|_2^2
\right],
\label{eq:flow_matching}
\end{equation}
where $\theta$ denotes the frozen pretrained parameters and $\psi$ denotes the trainable MeteoMoE and LoRA parameters.
To provide additional reconstruction-level supervision, we recover the clean latent estimate $\hat{z}_0$ from $z_t$ and the predicted velocity and decode it into $\hat{V}_{1:T}$.
We apply frame-wise reconstruction and perceptual losses~\cite{johnson2016perceptual}, \emph{i.e.},
\begin{equation}
\mathcal{L}_{\ell_1}
=
\frac{1}{T}
\sum_{i=1}^{T}
\left\|
\hat{V}_{i}
-
V_{i}
\right\|_1,
\qquad
\mathcal{L}_{\mathrm{vgg}}
=
\frac{1}{T}
\sum_{i=1}^{T}
\mathrm{VGG}
\left(
\hat{V}_{i},
V_{i}
\right).
\label{eq:reconstruction_losses}
\end{equation}
The overall training objective can be written as,
\begin{equation}
\mathcal{L}
=
\mathcal{L}_{\mathrm{fm}}
+
\lambda_{\ell_1}
\mathcal{L}_{\ell_1}
+
\lambda_{\mathrm{vgg}}
\mathcal{L}_{\mathrm{vgg}},
\label{eq:overall_loss}
\end{equation}
where $\lambda_{\ell_1}$ and $\lambda_{\mathrm{vgg}}$ are set to 0.1 and 0.05, respectively.

\begin{table*}[t!]
\centering
\caption{
Quantitative comparison across \textit{Weather Preservation}, \textit{Weather Introduction}, and \textit{Weather Removal}.
Overall Score is reported only for Weather Preservation, as intentional weather changes in Weather Introduction and Weather Removal make the aggregate input-consistency score less appropriate.
The best and second-best results are highlighted in \textbf{bold} and \underline{underlined}, respectively.
}
\label{tab:weather_capabilities}

\scriptsize
\setlength{\tabcolsep}{2.2pt}
\renewcommand{\arraystretch}{1.02}

\resizebox{\textwidth}{!}{
\begin{tabular}{@{}lccccccc@{\hspace{7pt}}cc@{\hspace{7pt}}ccc@{}}
\toprule
\multirow{2}{*}{Method}
& \multicolumn{7}{c}{VBench-I2V Evaluation}
& \multicolumn{2}{c}{Camera Evaluation}
& \multicolumn{3}{c}{Weather Evaluation}
\\
\cmidrule(lr){2-8}
\cmidrule(lr){9-10}
\cmidrule(lr){11-13}
& \shortstack{Overall\\Score$\uparrow$}
& \shortstack{Subject\\Consistency$\uparrow$}
& \shortstack{Background\\Consistency$\uparrow$}
& \shortstack{Motion\\Smoothness$\uparrow$}
& \shortstack{Dynamic\\Degree$\uparrow$}
& \shortstack{Aesthetic\\Quality$\uparrow$}
& \shortstack{Imaging\\Quality$\uparrow$}
& RotErr$\downarrow$
& TransErr$\downarrow$
& \shortstack{Weather\\Alignment$\uparrow$}
& \shortstack{VLM\\Evaluation$\uparrow$}
& \shortstack{User\\Study$\uparrow$}
\\

\midrule
\multicolumn{13}{c}{\textbf{Weather Preservation}}\\
\addlinespace[2pt]

Uni3C
& 86.53
& \underline{95.80}
& \underline{94.15}
& \underline{99.21}
& 30.50
& 50.87
& \underline{63.86}
& 0.755
& 0.102
& 88.00
& 74.99
& 76.80
\\

GEN3C
& 86.22
& \textbf{96.89}
& \textbf{94.34}
& \textbf{99.50}
& 23.50
& 50.11
& 60.46
& \textbf{0.331}
& \underline{0.073}
& 85.50
& 72.30
& 74.90
\\

VerseCrafter
& 85.01
& 95.20
& 92.68
& 99.17
& 23.00
& 50.86
& \textbf{66.21}
& 5.926
& 0.128
& \underline{90.75}
& \textbf{77.02}
& \underline{78.60}
\\

NeoVerse
& 85.73
& 92.83
& 92.15
& 99.15
& \underline{38.00}
& 48.76
& 61.93
& \underline{0.678}
& \textbf{0.054}
& 85.75
& 63.38
& 68.70
\\

LingBot-World
& \underline{86.56}
& 95.57
& 93.84
& 99.06
& \underline{38.00}
& \textbf{51.44}
& 63.45
& 4.225
& 0.173
& 88.76
& 74.43
& 75.80
\\

Ours
& \textbf{86.70}
& 94.78
& 94.03
& 98.86
& \textbf{43.50}
& \underline{50.93}
& 63.66
& 3.416
& 0.132
& \textbf{93.00}
& \underline{76.07}
& \textbf{80.90}
\\

\midrule
\multicolumn{13}{c}{\textbf{Weather Introduction}}\\
\addlinespace[2pt]

Uni3C
& --
& 95.37
& \underline{94.26}
& 99.17
& 28.00
& 50.09
& \underline{64.46}
& 0.796
& 0.161
& 12.00
& 22.50
& 28.60
\\

GEN3C
& --
& \textbf{96.84}
& \textbf{94.55}
& \textbf{99.49}
& 23.00
& 49.15
& 60.90
& \textbf{0.364}
& \textbf{0.056}
& 5.00
& 14.76
& 21.40
\\

VerseCrafter
& --
& 94.85
& 92.48
& 99.11
& 19.00
& 50.10
& \textbf{67.46}
& 5.991
& 0.134
& \underline{29.00}
& \underline{33.70}
& \underline{38.20}
\\

NeoVerse
& --
& 92.85
& 91.84
& \underline{99.21}
& \underline{39.00}
& 47.98
& 61.06
& \underline{0.679}
& \underline{0.068}
& 9.00
& 17.98
& 24.10
\\

LingBot-World
& --
& \underline{95.61}
& 93.55
& 99.10
& 35.00
& \underline{50.49}
& 64.05
& 4.316
& 0.155
& 13.00
& 20.13
& 26.70
\\

Ours
& --
& 95.28
& 94.15
& 98.90
& \textbf{47.00}
& \textbf{51.09}
& 61.51
& 3.207
& 0.122
& \textbf{61.00}
& \textbf{54.15}
& \textbf{66.70}
\\

\midrule
\multicolumn{13}{c}{\textbf{Weather Removal}}\\
\addlinespace[2pt]

Uni3C
& --
& 95.31
& 93.78
& \underline{99.25}
& 35.00
& \underline{51.82}
& 63.70
& 0.828
& 0.130
& 31.00
& 45.03
& 49.20
\\

GEN3C
& --
& \textbf{96.94}
& \underline{94.12}
& \textbf{99.51}
& 24.00
& 51.02
& 60.08
& \textbf{0.327}
& \underline{0.082}
& 34.00
& 46.17
& 50.70
\\

VerseCrafter
& --
& 94.25
& 92.77
& 99.22
& 30.00
& 51.31
& \underline{65.68}
& 6.784
& 0.122
& 31.00
& 41.67
& 46.80
\\

NeoVerse
& --
& 92.66
& 92.43
& 99.07
& 37.00
& 49.47
& 62.55
& \underline{0.671}
& \textbf{0.047}
& \underline{38.00}
& 40.98
& 45.60
\\

LingBot-World
& --
& 95.08
& 93.72
& 99.05
& \underline{44.00}
& \textbf{52.15}
& 63.54
& 4.384
& 0.303
& 34.00
& \underline{46.45}
& \underline{51.40}
\\

Ours
& --
& \underline{95.66}
& \textbf{94.14}
& 99.00
& \textbf{48.00}
& 50.95
& \textbf{67.03}
& 3.476
& 0.110
& \textbf{89.00}
& \textbf{77.92}
& \textbf{82.40}
\\

\bottomrule
\end{tabular}
}
\vspace{-4mm}
\end{table*}

\section{Experiments}

\subsection{Implementation Details}

We adopt a progressive training strategy that increases the clip length from 21 to 41 and finally 81 frames, while increasing the spatial resolution from $240\times416$ to $480\times832$.
Training is performed on 8 NVIDIA A800-SXM4 GPUs with a per-GPU batch size of 1 using BF16 precision.
We use AdamW with cosine learning-rate decay and gradient clipping at 1.0.
The learning rate is set to $1\times10^{-5}$ for the LoRA parameters and $5\times10^{-5}$ for the remaining trainable parameters.
The weather-state predictor is LoRA-fine-tuned using AdamW with a learning rate of $1\times10^{-4}$ and a per-device batch size of 4.
Greedy decoding is used for weather-state prediction at inference.

\subsection{Evaluation Configurations}

We evaluate on 100 held-out real-world scenes under four weather settings, \emph{i.e.}, sunny-to-sunny, sunny-to-adverse, adverse-to-sunny, and adverse-to-adverse, resulting in 400 test cases.
Sunny-to-sunny and adverse-to-adverse constitute \textit{Weather Preservation}, while sunny-to-adverse and adverse-to-sunny correspond to \textit{Weather Introduction} and \textit{Weather Removal}, respectively.
Weather Preservation is equally averaged over its sunny and adverse-weather subsets.
All methods are evaluated using the same input images, scene descriptions, target-weather instructions, and camera trajectories.
We evaluate video quality using VBench-I2V~\cite{huang2025vbench++}.
Camera-control performance is measured using rotation and translation errors estimated by VGGT-Omega~\cite{wang2026vggt}.
Weather controllability is evaluated using Weather Alignment, VLM Evaluation, and User Study.
Overall Score is reported only for Weather Preservation, as intentional weather changes in Weather Introduction and Weather Removal make the aggregate input-consistency score less appropriate.
Detailed metric definitions, generation settings, and the user-study protocol are in Appendix~\ref{app:evaluation_details}.

\subsection{Comparison with State-of-the-Art Methods}

We compare MeteoVerse with five state-of-the-art controllable video world models with publicly available implementations or checkpoints, including Uni3C~\cite{cao2025uni3c}, GEN3C~\cite{ren2025gen3c}, VerseCrafter~\cite{zheng2026versecrafter}, NeoVerse~\cite{yang2026neoverse}, and LingBot-World~\cite{robbyant2026lingbotworld}.
All methods are evaluated under the same protocol.

\noindent\textbf{Quantitative comparisons.}
The quantitative results are reported in Tab.~\ref{tab:weather_capabilities}.
Under Weather Preservation, MeteoVerse achieves the highest Overall Score, Weather Alignment, and User Study score, while remaining competitive on the individual VBench-I2V and camera-control metrics.
The advantage becomes much clearer when the weather condition needs to change.
For Weather Introduction, MeteoVerse improves Weather Alignment from $29.00$ to $61.00$, VLM Evaluation from $33.70$ to $54.15$, and User Study from $38.20$ to $66.70$.
For Weather Removal, the corresponding scores improve from $38.00$ to $89.00$, $46.45$ to $77.92$, and $51.40$ to $82.40$.
Although several baselines perform better on individual video-quality or camera metrics, these metrics mainly reflect scene consistency and trajectory following and can remain high even when the requested weather change is not realized.
By explicitly modeling the weather transition, MeteoVerse achieves substantially stronger weather controllability while retaining competitive overall generation quality.

\begin{figure*}[t!]
    \centering
    \includegraphics[width=\textwidth]{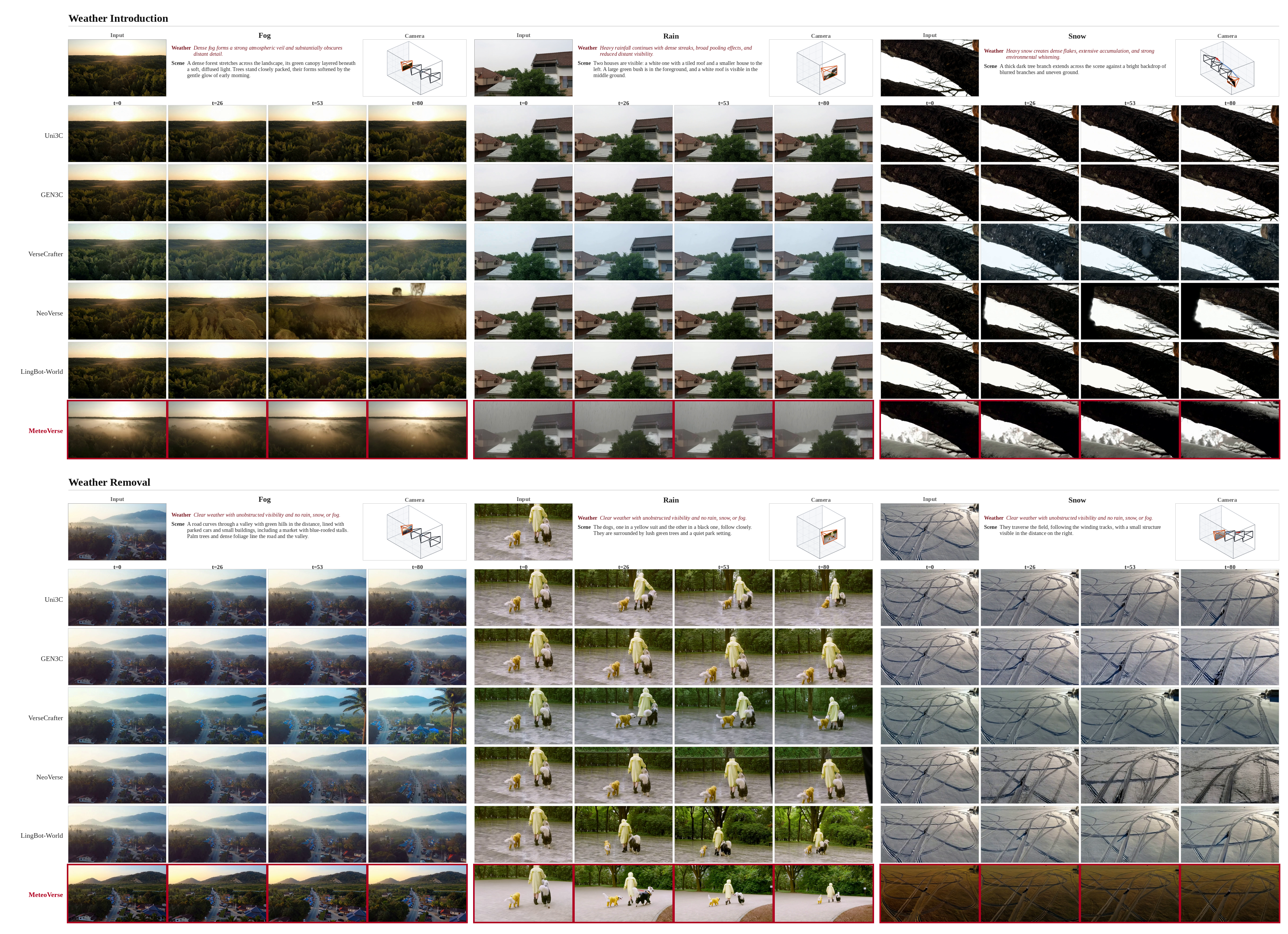}
    \caption{
    Visual comparisons on weather introduction and removal.
    More results are in the \textit{Suppl}.
    }
    \label{fig:comparisons}
    \vspace{-2mm}
\end{figure*}

\noindent\textbf{Qualitative comparisons.}
Fig.~\ref{fig:comparisons} presents qualitative comparisons on Weather Introduction and Weather Removal.
For Weather Introduction, existing world models often retain the original weather appearance or produce only weak target-weather effects, whereas MeteoVerse more clearly realizes the requested rain, snow, and fog conditions.
For Weather Removal, competing methods frequently leave visible adverse-weather effects, while MeteoVerse more effectively suppresses them and produces a cleaner scene appearance.
The underlying scene content and camera-driven evolution remain visually consistent during weather manipulation.
These qualitative observations are consistent with the quantitative weather-control results.
Video comparisons are provided in the \textit{Suppl}.

\section{Ablation Studies}
\label{sec:ablation}

\begin{table*}[t!]
\centering
\caption{
Ablation on weather-control representations.
All metrics are equally averaged over Weather Preservation, Weather Introduction, and Weather Removal.
}
\label{tab:ablation_weather_transition}
\vspace{-2mm}

\scriptsize
\setlength{\tabcolsep}{2.4pt}
\renewcommand{\arraystretch}{1.02}

\resizebox{\textwidth}{!}{
\begin{tabular}{@{}lcccccc@{\hspace{7pt}}cc@{\hspace{7pt}}ccc@{}}
\toprule
\multirow{2}{*}{Methods}
& \multicolumn{6}{c}{VBench-I2V Evaluation}
& \multicolumn{2}{c}{Camera Evaluation}
& \multicolumn{3}{c}{Weather Evaluation}
\\
\cmidrule(lr){2-7}
\cmidrule(lr){8-9}
\cmidrule(lr){10-12}
& \shortstack{Subject\\Consistency$\uparrow$}
& \shortstack{Background\\Consistency$\uparrow$}
& \shortstack{Motion\\Smoothness$\uparrow$}
& \shortstack{Dynamic\\Degree$\uparrow$}
& \shortstack{Aesthetic\\Quality$\uparrow$}
& \shortstack{Imaging\\Quality$\uparrow$}
& RotErr$\downarrow$
& TransErr$\downarrow$
& \shortstack{Weather\\Alignment$\uparrow$}
& \shortstack{VLM\\Evaluation$\uparrow$}
& \shortstack{User\\Study$\uparrow$}
\\
\midrule

Target-weather Prompt $p_{\mathrm{w}}$
& 92.60 & 93.14 & 98.87 & 48.50 & 51.88 & 61.30
& 4.084 & \textbf{0.115} & 80.85 & 63.98 & 73.63
\\

Target State $\mathbf{s}_{\mathrm{tar}}$
& 93.96 & 93.29 & 98.78 & \textbf{58.33} & \textbf{51.92} & 61.76
& 4.372 & 0.189 & 66.03 & 58.73 & 64.07
\\

Transition State $\mathbf{s}_{\mathrm{trans}}$ (Ours)
& \textbf{95.24} & \textbf{94.11} & \textbf{98.92} & 46.17 & 50.99 & \textbf{64.07}
& \textbf{3.366} & 0.121 & \textbf{81.00} & \textbf{69.38} & \textbf{76.67}
\\

\bottomrule
\end{tabular}
}
\vspace{-4mm}
\end{table*}

\begin{figure*}[t!]
    \centering
    \includegraphics[
        width=0.99\textwidth,
        trim=0 2 0 0,
        clip
    ]{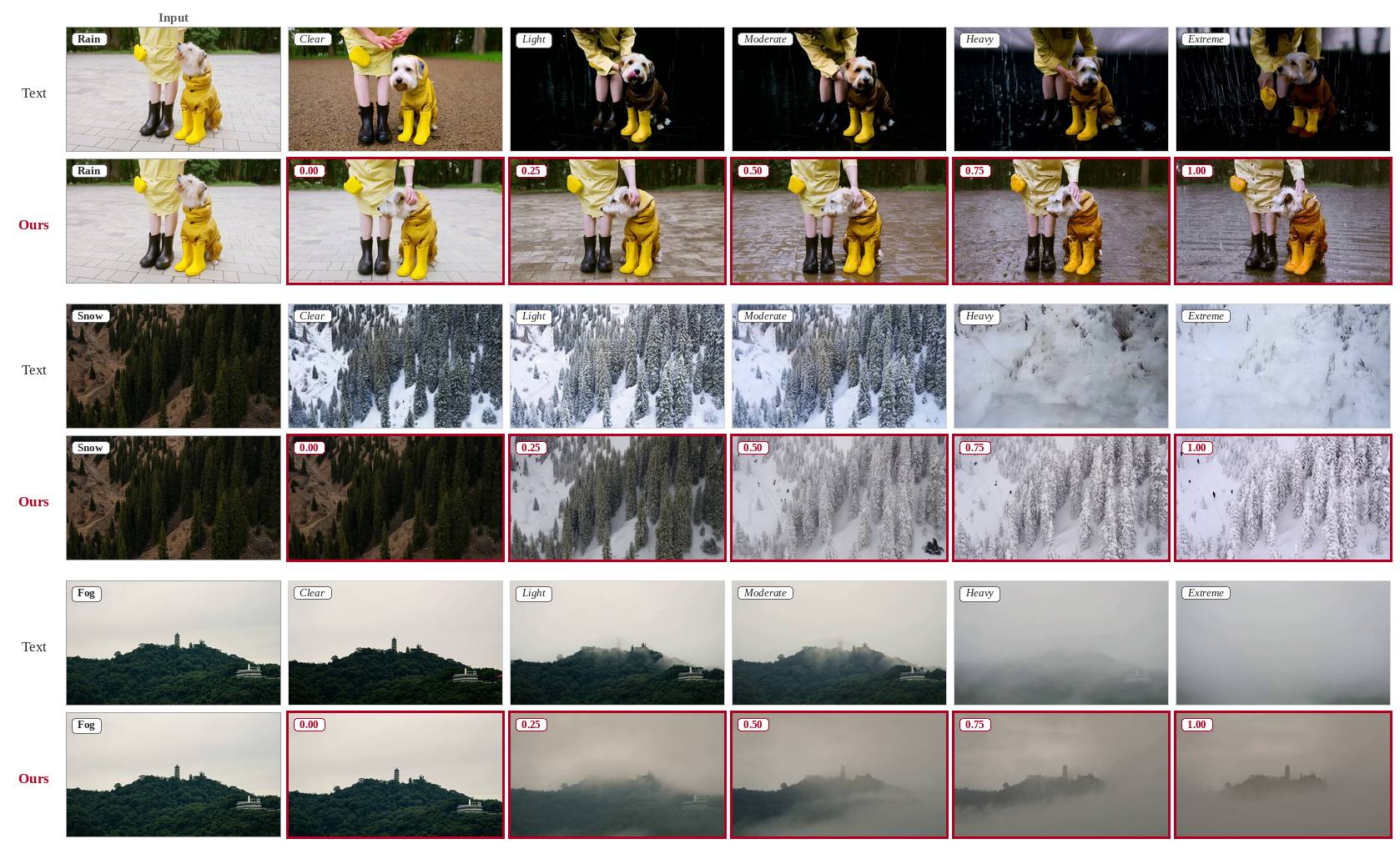}
    \caption{
    Visual results of fine-grained weather control. More results are in the \textit{Suppl}.
    }
    \label{fig:fine_grained_intensity_control}
    \vspace{-4mm}
\end{figure*}

\begin{table*}[t!]
\centering
\caption{
Ablation on MeteoMoE and backbone adaptation.
All metrics are equally averaged over Weather Preservation, Weather Introduction, and Weather Removal.
}
\label{tab:ablation_meteomoe}
\vspace{-2mm}

\scriptsize
\setlength{\tabcolsep}{2.4pt}
\renewcommand{\arraystretch}{1.02}

\resizebox{\textwidth}{!}{
\begin{tabular}{@{}lcccccc@{\hspace{7pt}}cc@{\hspace{7pt}}ccc@{}}
\toprule
\multirow{2}{*}{Methods}
& \multicolumn{6}{c}{VBench-I2V Evaluation}
& \multicolumn{2}{c}{Camera Evaluation}
& \multicolumn{3}{c}{Weather Evaluation}
\\
\cmidrule(lr){2-7}
\cmidrule(lr){8-9}
\cmidrule(lr){10-12}
& \shortstack{Subject\\Consistency$\uparrow$}
& \shortstack{Background\\Consistency$\uparrow$}
& \shortstack{Motion\\Smoothness$\uparrow$}
& \shortstack{Dynamic\\Degree$\uparrow$}
& \shortstack{Aesthetic\\Quality$\uparrow$}
& \shortstack{Imaging\\Quality$\uparrow$}
& RotErr$\downarrow$
& TransErr$\downarrow$
& \shortstack{Weather\\Alignment$\uparrow$}
& \shortstack{VLM\\Evaluation$\uparrow$}
& \shortstack{User\\Study$\uparrow$}
\\
\midrule

Global Transition Conditioning
& 93.53 & 92.98 & 98.76 & \textbf{64.17} & \textbf{51.90} & \textbf{67.51}
& 4.291 & \textbf{0.095} & 58.33 & 54.75 & 58.42
\\

Shared Weather Expert
& 93.60 & 93.09 & 98.80 & 62.33 & 51.69 & 66.30
& 4.370 & 0.114 & 65.00 & 58.96 & 63.74
\\

w/o Backbone LoRA
& 94.31 & 93.56 & 98.90 & 45.50 & 51.60 & 66.83
& 4.396 & 0.141 & 52.02 & 51.55 & 53.82
\\

Full MeteoMoE (Ours)
& \textbf{95.24} & \textbf{94.11} & \textbf{98.92} & 46.17 & 50.99 & 64.07
& \textbf{3.366} & 0.121 & \textbf{81.00} & \textbf{69.38} & \textbf{76.67}
\\

\bottomrule
\end{tabular}
}
\vspace{-4mm}
\end{table*}

\subsection{Effect of Weather Transition Modeling}
\label{subsec:ablation_weather_transition}

We compare the target-weather prompt $p_{\mathrm{w}}$, the target weather state $\mathbf{s}_{\mathrm{tar}}$, and the transition state $\mathbf{s}_{\mathrm{trans}}$.
The target state describes the desired weather, whereas the transition state additionally accounts for the observed condition and directly specifies the required weather modification.
As shown in Tab.~\ref{tab:ablation_weather_transition}, $\mathbf{s}_{\mathrm{trans}}$ improves Weather Alignment, VLM Evaluation, and User Study from $66.03$, $58.73$, and $64.07$ with $\mathbf{s}_{\mathrm{tar}}$ to $81.00$, $69.38$, and $76.67$.
Compared with $p_{\mathrm{w}}$, it achieves similar Weather Alignment but higher VLM Evaluation and User Study, showing the benefit of explicitly representing the required weather change.
We further evaluate fine-grained intensity control for weather introduction by fixing a sunny input, scene description, camera trajectory, and random seed while varying only the target weather intensity.
As shown in Fig.~\ref{fig:fine_grained_intensity_control}, explicit transition control produces a clearer and more gradual progression of introduced rain, snow, and fog than text-based control.
Video demonstrations are in the \textit{Suppl}.

\subsection{Effect of MeteoMoE Architecture}
\label{subsec:ablation_meteomoe}

We evaluate how the weather transition is incorporated into the video backbone.
\emph{Global Transition Conditioning} projects $\mathbf{s}_{\mathrm{trans}}$ into a global embedding and injects it directly into each DiT block, while \emph{Shared Weather Expert} replaces the category-specific rain, snow, and fog experts with a single shared expert.
As shown in Tab.~\ref{tab:ablation_meteomoe}, Weather Alignment improves from $58.33$ with global conditioning and $65.00$ with the shared expert to $81.00$ with the full MeteoMoE, with consistent gains in VLM Evaluation and User Study.
These results demonstrate the benefit of category-specific weather experts.
Removing the backbone LoRA adapters further reduces Weather Alignment to $52.02$, showing that lightweight backbone adaptation is important for effectively incorporating the transition-aware features.
The full MeteoMoE also achieves the highest subject consistency, background consistency, and motion smoothness.
More ablation results about the  weather-state predictor are provided in Appendix~\ref{app:effect_weather_predictor}.

\section{Conclusion}
\label{sec:conclusion}

We presented MeteoVerse, a unified weather-controllable video world model for camera-controlled future prediction from a single sunny or adverse-weather image.
MeteoVerse explicitly models the required weather transition and uses MeteoMoE to realize weather preservation, introduction, and removal, with fine-grained intensity control for weather introduction.
We further constructed the MeteoVerse dataset with over 50K real-world weather clips, generated sunny counterparts, disentangled scene and weather descriptions, weather-intensity annotations, and camera trajectories.
Extensive experiments demonstrate substantially improved weather controllability while retaining competitive scene consistency, temporal coherence, and camera-control performance.

\bibliography{references}
\bibliographystyle{iclr2027_conference}

\appendix

\section{Additional Evaluation Details}
\label{app:evaluation_details}

\noindent\textbf{Benchmark.}
The held-out benchmark contains 100 real-world scenes with no clip- or source-level overlap with the training set.
Each scene is evaluated under sunny-to-sunny, sunny-to-adverse, adverse-to-sunny, and adverse-to-adverse settings, yielding 400 test cases.
Weather Preservation is averaged equally over its sunny and adverse-weather subsets.
MeteoVerse generates 81-frame videos at $480\times832$ using 40 denoising steps, with classifier-free guidance set to 5.0.

\noindent\textbf{Evaluation metrics.}
We evaluate three complementary aspects.
Video quality is measured using the VBench-I2V~\cite{huang2025vbench++} metrics, including subject consistency, background consistency, motion smoothness, dynamic degree, aesthetic quality, and imaging quality.
Overall Score is reported only for Weather Preservation, as intentional weather changes in Weather Introduction and Weather Removal make the aggregate input-consistency score less appropriate.
Camera-control performance is evaluated using camera trajectories estimated from all generated frames by VGGT-Omega~\cite{wang2026vggt}.
We report geodesic rotation error (RotErr) and scale-aligned translation error (TransErr).
Weather controllability is evaluated using Weather Alignment and VLM Evaluation.
Qwen~3.8~Max~\cite{teamqwen3} receives only the target-weather instruction and generated video.
Weather Alignment measures the percentage of samples that exhibit the requested weather condition, while VLM Evaluation measures the overall quality of weather realization using a 0--100 score over weather-scene compatibility, weather dynamics, and overall video quality.
For the ablation studies, all metrics are averaged over Weather Preservation, Weather Introduction, and Weather Removal.

\noindent\textbf{User study.}
We conduct a blind user study on 36 benchmark cases, with 12 cases for each of Weather Preservation, Weather Introduction, and Weather Removal.
The preservation subset contains equal numbers of sunny- and adverse-weather cases, while the introduction and removal subsets are balanced across rain, snow, and fog.
We recruit 24 participants.
All videos are anonymized and presented in randomized order.
Each participant evaluates a randomly assigned subset, and each generated result receives at least five independent ratings.
Given the input image, target-weather instruction, and generated video, participants rate weather faithfulness, preservation of non-weather scene content, and temporal naturalness on a five-point scale.
For each rating, the three scores are equally averaged and linearly mapped to $[0,100]$.
We then average the scores across participants and test cases.
For the main comparison, we report a separate User Study score for each weather-control capability.
For the ablation and predictor analyses, we report the equal-weight macro-average over Weather Preservation, Weather Introduction, and Weather Removal.

\section{Effect of Weather-State Predictor}
\label{app:effect_weather_predictor}

We evaluate the weather-state predictor on 400 test cases from 100 held-out scenes.
It achieves $99.00\%$ transition accuracy, measuring whether weather preservation, introduction, or removal is correctly identified.
For continuous state estimation, it obtains an intensity MAE of $0.0342$ and a state MAE of $0.0119$.
Under a maximum component-wise error tolerance of $0.10$, the state accuracy reaches $89.75\%$.
We further examine the impact of prediction errors on downstream generation by replacing ground-truth weather states with predicted states in the same frozen MeteoVerse model.
As shown in Tab.~\ref{tab:weather_predictor_world_model}, Weather Alignment changes from $82.00$ to $81.00$, while VLM Evaluation and User Study change from $70.18$ and $77.56$ to $69.38$ and $76.67$, respectively.
The small differences indicate that the predicted weather states provide conditioning close to the ground-truth states for downstream weather control.

\begin{table*}[t!]
\centering
\caption{
Effect of weather-state prediction on downstream generation.
All metrics are equally averaged over Weather Preservation, Weather Introduction, and Weather Removal.
}
\label{tab:weather_predictor_world_model}

\scriptsize
\setlength{\tabcolsep}{2.4pt}
\renewcommand{\arraystretch}{1.02}

\resizebox{\textwidth}{!}{
\begin{tabular}{@{}lcccccc@{\hspace{7pt}}cc@{\hspace{7pt}}ccc@{}}
\toprule
\multirow{2}{*}{Methods}
& \multicolumn{6}{c}{VBench-I2V Evaluation}
& \multicolumn{2}{c}{Camera Evaluation}
& \multicolumn{3}{c}{Weather Evaluation}
\\
\cmidrule(lr){2-7}
\cmidrule(lr){8-9}
\cmidrule(lr){10-12}
& \shortstack{Subject\\Consistency$\uparrow$}
& \shortstack{Background\\Consistency$\uparrow$}
& \shortstack{Motion\\Smoothness$\uparrow$}
& \shortstack{Dynamic\\Degree$\uparrow$}
& \shortstack{Aesthetic\\Quality$\uparrow$}
& \shortstack{Imaging\\Quality$\uparrow$}
& RotErr$\downarrow$
& TransErr$\downarrow$
& \shortstack{Weather\\Alignment$\uparrow$}
& \shortstack{VLM\\Evaluation$\uparrow$}
& \shortstack{User\\Study$\uparrow$}
\\
\midrule

GT States
& 94.35 & 93.65 & 98.87 & \textbf{52.33} & 50.68 & 63.26
& 3.695 & \textbf{0.120} & \textbf{82.00} & \textbf{70.18} & \textbf{77.56}
\\

Predicted States (Ours)
& \textbf{95.24} & \textbf{94.11} & \textbf{98.92} & 46.17 & \textbf{50.99} & \textbf{64.07}
& \textbf{3.366} & 0.121 & 81.00 & 69.38 & 76.67
\\

\bottomrule
\end{tabular}
}
\end{table*}

\end{document}